\documentclass[conference]{IEEEtran}
\IEEEoverridecommandlockouts                              
\usepackage{graphics}
\usepackage{cite}
\usepackage{float}
\usepackage{caption}
\usepackage{multicol}
\usepackage{multirow}
\usepackage{graphicx}
\usepackage{amsmath}
\usepackage{hyperref}

\usepackage{enumitem,kantlipsum}
\graphicspath{images/}
\usepackage{tabularx}
\title{Accurate Crop Spraying with RTK and Machine Learning on an Autonomous Field Robot}
\author{W.~M.~T.~D.~Wijesundara, T.~D.~Wanigathunga, M.~N.~C.~Waas, R.~T.~Hithanadura, \\S.~R.~Munasinghe,~\IEEEmembership{Senior Member,~IEEE}
\thanks{All authors are from the Department of Electronic and Telecommunication Engineering, University of Moratuwa, Sri Lanka}
\thanks{Prof. S. R. Munasinghe is a visiting Fellow at the Dept of Global Development, College of Agriculture and Life Sciences, Cornell University, NY, USA}}
\begin{document}
\maketitle
\begin{abstract}
	The agriculture sector requires a lot of labor and resources. Hence, the farmers are constantly being pressed for technology and automation to be cost-effective. In this context, autonomous robots can play a very in very important role in carrying out agricultural tasks such as spraying, sowing, inspection, and even harvesting. This paper presents one such autonomous robot that is able to identify plants and spray agro-chemicals precisely. The robot uses machine vision technologies to find plants and RTK-GPS technology to navigate the robot along a predetermined path. The experiments were conducted in a field of potted plants in which successful results have been obtained. 
\end{abstract}
\section{Introduction}
	In a variety of industries, autonomous robots are used to streamline routine and repetitive tasks. Recent years have seen an increase in the use of autonomous robots in the agricultural sector for a variety of tasks, including watering, spraying, seeding, harvesting, etc. Furthermore, autonomous robots are used in crop fields to monitor various tasks \cite{kusumam20173d}, \cite{nakarmi2014within} and also for weed control \cite{mccool2018efficacy}, \cite{wu2019design}. This approach is useful in fields where the crops are planted in parallel rows. To carry out desired tasks, robots must be able to navigate through these rows accurately. This research project demonstrates a spraying robot that uses RTK-GPS \cite{rtk_link} for navigation and machine vision for plant detection.\par
	The RTK technology is popular in the localization and navigation of field robots and vehicles. This technology increases the position accuracy over the conventional GPS (Global Positioning System) technology. With the RTK technology, it is possible to localize and guide the robot along a predefined path with centimeter-level precision.\par
	One of the project's primary considerations is robot localization and navigation. The robot needs to travel along straight lines formed by a set of pre-defined locations such as the start and the end of a passage between adjacent plant rows. The GPS path can be created by fetching these locations to the mission planner \cite{misson_planner}. A combination of sensors: RTK, wheel encoders, and an inertial measurement unit (IMU) is used to estimate the position in real time while the robot is moving. The sensor fusion method combines multiple sensors and produces the best estimate. Due to the presence of RTK, the estimated position is expected to be accurate enough. The following sections will provide a detailed explanation of the internal algorithm used to keep the robot moving along the planned path as closely as possible.
	With the aid of machine learning, the robot can recognize plants and decide how to treat them appropriately. A customized treatment process is conceivable with the help of advanced machine vision technologies going beyond the basic watering process. Spraying can be made cost-effective due to automated plant detection and targeted spraying. Machine learning and image processing are used to recognize and track the plants using spherical coordinates while the robot is moving. The ability to spray the optimal amount of fertilizer onto the right spots helps to reduce the over-usage of agro-chemicals.\par
	After recognizing a plant in the camera image, the position coordinates of the plant are calculated. Then, the plant is localized with respect to the spray nozzle using co-ordinate transformation from the camera to the spray nozzle. This process keeps tracking the plant as the robot moves, and the two nozzles spray agro-chemicals right onto the plant. The Robot Operating System (ROS) running on a Jetson Nano single board computer was chosen as the development platform.\par
	A variety of technologies have been used with autonomous systems in the field of agriculture. In \cite{Olson2011AprilTagAR} and \cite{88147}, navigation in crop fields using visual markers have been presented, where Bell \cite{bell} and  Thuilot \textit{et al}. \cite{thuilot} have used GNSS technology to guide the robot through the crop fields. Multiple sensors, including cameras, laser scanners, GNSS receivers, and others have been used for more precise agricultural tasks \cite{Underwood2015RealtimeTD},\cite{Imperoli2018AnEM}. Dong \textit{et al}. \cite{Dong20174DCM}, and Chebrolu \textit{et al}. \cite{Chebrolu2019RobotLB} used a number of sensors in conjunction with previously mapped field data to locate and steer the robot in a crop field. However, they are either not very accurate in the aforementioned scenarios or they used numerous sensors to make them accurate. In this research, the robot uses RTK technology in conjunction with a number of other sensors and ensures accurate localization and navigation.
	In \cite{Billingsley1997TheSD} and \cite{astrand}, localization and navigation are based on a single-position sensor. Hence, the accuracy and functionality are affected by the quality of the measurements and the real-time availability of the sensor measurement. In \cite{143350} visual-servoing technique, which is typically used to control robot manipulators has been tested to find row crop fields. The methods presented in Cherubini \textit{et al}. \cite{Cherubini2008AnIV}, \cite{768184}, \cite{this} describe how to steer the robot along continuous paths. In this research, precise RTK data is used together with vision data in order to maintain high accuracy in localization, navigation, and most importantly, in crop spraying.
\section{System Overview}
	This study concerns small to moderate-sized plants that are in a structured arrangement. The spraying robot must move along a pre-planned path while detecting and spraying plants.
\subsection{Robot Design}
\begin{figure}[h]
	\centering
	\includegraphics[width=0.45\textwidth]{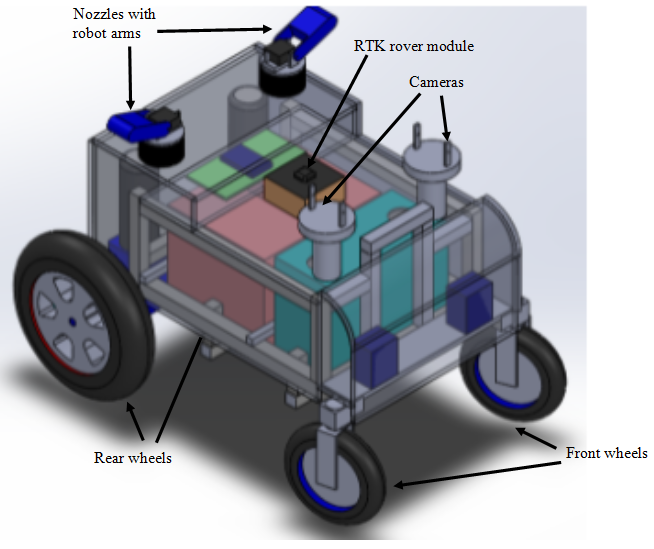}
	\caption{Robot design}
	\label{entire-design}
	\end{figure}

	The spray robot is shown in Fig.\ref{entire-design}
	There are three major hardware layers in the robot platform design. The batteries and agro-chemical tank are located in the bottom layer. The capacity of the tank is 10ltrs and is located in a 29cm$\times$14cm$\times$24cm space in the bottom later. Forty-five plants can be sprayed with one full tank assuming 200 ml of liquid per plant. The tank and the pressure pump are located at the rear part of the robot closer to the rear wheels creating enough separation from the electronic system. The central controller is located in the middle layer. This is a printed circuit board including the robot motor drives, water pump control, the main processor (a Jetson nano), and a low-level robot-controlling microcontroller (an Arduino mega). The top layer consists of parts that are used to interact with the user and the environment. It has two cameras mounted on camera holders, two nozzles mounted on two robot arms, an RTK (reach M+) module and its antenna, a buzzer, LED indicators, a liquid crystal display, and all the switches needed to interact with the user. In order to shield other parts of the robot from fertilizer, the two nozzles are placed at the farther back end of the robot. Additionally, some parts, like two cameras and a reach M+ module, have an additional shield to protect them. The robot platform design is shown in Fig. \ref{entire-design}.
\subsection{Autonomous Crop Spraying Process}
\begin{figure}[H]
	\centering
	\includegraphics[width=0.45\textwidth]{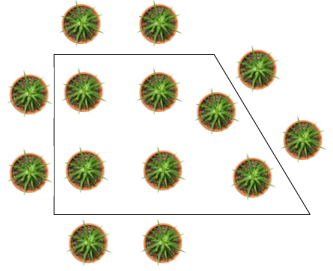}
	\caption{A candidate path for plant spraying}
	\label{path}
	\end{figure}

	First, the path is planned as shown in Fig.\ref{path} through the plants. The shortest path assuring every plant is attended to is planned. Yet, the optimum path planning is not addressed in this research. The two cameras mounted on the robot provide a video stream to the vision processing system which detects the plants. Once detected, the distance and direction to a selected plant are determined and sent to the spray control system to control the robot arms, pressure pump, and nozzles.
	\subsection{Robot System Design}
\begin{figure}[H]
	\centering
	\includegraphics[width=0.45\textwidth]{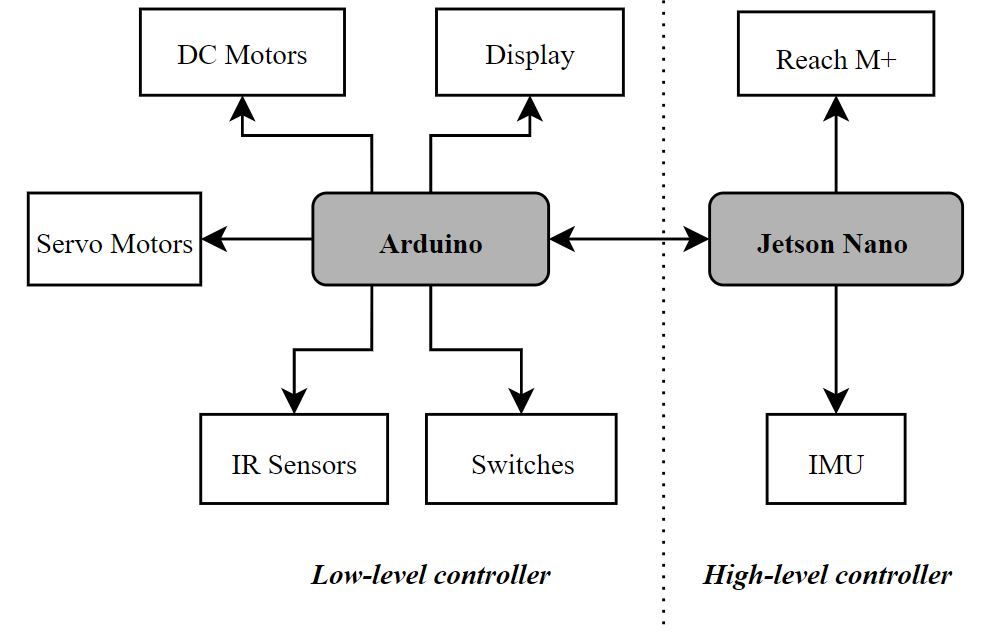}
	\caption{Robot system design}
	\label{system}
	\end{figure}

	The system design is shown in the Figure. \ref{system}.
	Utilizing two different levels of controllers allows for the distribution of control over low-level and high-level systems and components; a ROS-installed Jetson Nano as the high-level controller and an Arduino board as the low-level controller. The robot's low-level components comprise three IR sensors to monitor the obstacles, two geared motors to drive the wheels, two servo motors each for the two nozzles of the spray system, and a display. The high-level components are a Reach M+ to receive RTK GPS position, and an IMU to read the attitude of the robot.\par
	The system was implemented in ROS (robot operating system) that is running on the Jetson nano main control board. Each block shown is controlled by a ROS node. The main ROS node triggers the other nodes with data request scripts and implements the real-time operation. The low-level motor controlling node receives the appropriate signals from the navigation node and controls the motor speeds. Tasks such as obstacle detection and locating plants are handled by respective nodes and communicated to the main ROS node.
\section{System Architecture}
\begin{figure*}[h]
	\centering
	\includegraphics[width=1.0\textwidth]{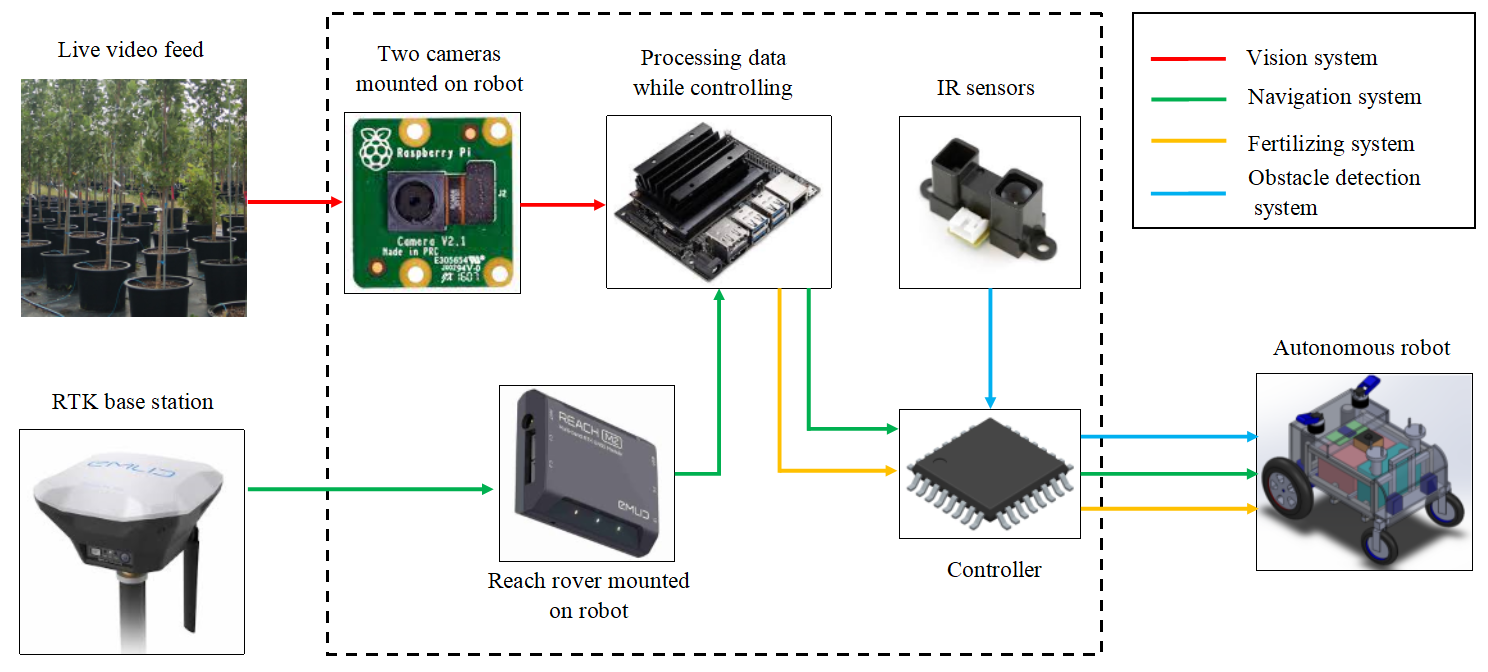}
	\caption{System Architecture}
	\label{architecture}
	\end{figure*}

	Figure \ref{architecture} shows the system architecture. The RTK positioning system, which is used for localization and navigation is the main component, and the two cameras are used to see the close vicinity of the robot. The cameras stream live video to the Jetson Nano single-board computer, which processes the images to identify the plants to be sprayed.
\subsection{RTK Positioning System}
	Two receiver modules are used in the RTK system- the base station (fixed), and the rover (fixed or moving). The base station is located at a position where the latitude and longitude coordinates are known. In this research, the base station was Emlid RS and the rover module was Emlid M+. Both modules were configured using Emlid software. The base station uses GNSS (Global Navigation Satellite System) signals and its known position to calculate the GNSS position error, which is transmitted to the rover over a 915MHz RF (Radio Frequency) link. The rover also receives GNSS signals and it uses the GNSS position error sent by the base station to accurately calculate its position.
	The RTK system communicates the robot's position to the Jetson nano central processor to localize and navigate the robot through the pre-specified path.
\subsection{Navigation System}
\begin{figure}[ht]
	\centering
	\includegraphics[width=0.45\textwidth]{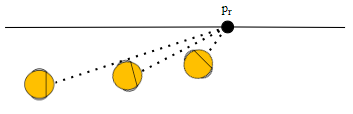}\\
	\includegraphics[width=0.45\textwidth]{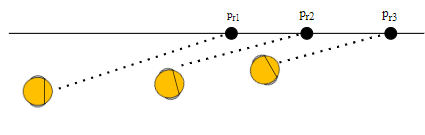}
	\caption{Top: Static reference point, Bottom: Dynamic reference point method}
	\label{static_dynamic}
	\end{figure}

\begin{figure}[ht]
	\centering
	\includegraphics[width=0.4\textwidth]{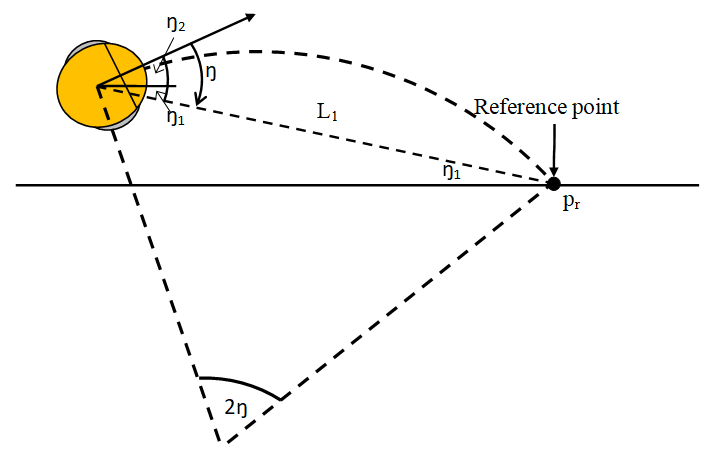}\\
	\includegraphics[width=0.4\textwidth]{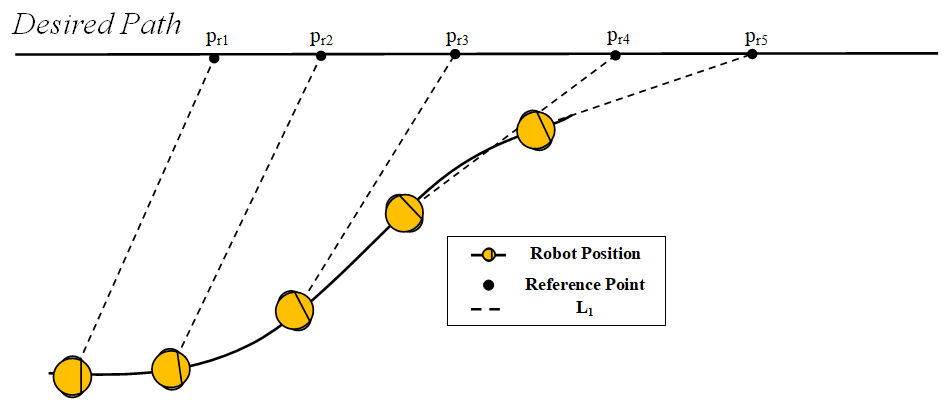}
	\caption{Top: Planning for correction, Bottom: Reaching the reference path}
	\label{approach}
	\end{figure}

	As the robot moves, it tends to deviate from the reference path, and corrective control actions are needed to pull the robot back to the reference path. Figure \ref{static_dynamic} illustrates the static reference and dynamic reference method as candidate control policies for path correction.
	In the static reference point method, the reference point is fixed on the reference path within a period of time, whereas in the dynamic reference point method, the reference point moves along the reference path as the robot follows it. In this research, the dynamic reference point method was used accepting the recommendation by Sanghyuk \textit{et al}. \cite{park2004new} where it shows that this method can bring the robot to the reference path and keep it there with minimum deviations. The dynamic reference point method works as follows: First, the line $L_1$ with an appropriate distance that is shown in Fig.\ref{approach} is decided. Then, the coordinates of the reference point $P$ and $\eta$ the angle between $L_1$ and the robot's heading are determined. A circular arc shown in Fig\ref{approach} goes through the robot along its heading, and the reference point $P$ is drawn. This arc is used to generate wheel speeds that will drive the robot along the arc. Successive applications of this control policy will eventually help keep the robot on the reference path as shown in Fig. \ref{approach} bottom.
	\subsection{Dead Reckoning for Self-Localization}
\begin{figure}[ht]
	\centering
	\includegraphics[width=0.45\textwidth]{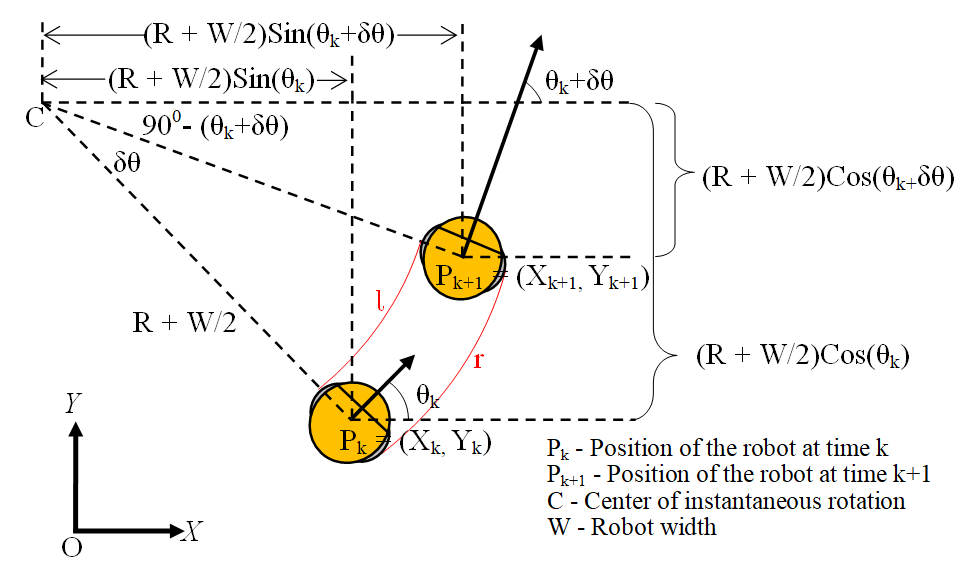}
	\caption{Position and heading difference over a small time interval}
	\label{odom}
	\end{figure}

	The robot's motion within a small time interval is shown in Fig. \ref{odom}.
	At time step $k$ the left and right wheels travel $l_k=R_k\delta\theta_k$ and $r_k=\delta\theta_k(R_k+w)$ distances. The heading changes $\delta\theta_k$ and the instantaneous radius of curvature $R_k$ of the motion are determined as follows.
	\begin{eqnarray}
		\delta\theta_k &=&\frac{r_k-l_k}{w}\\
		R_k &=&\frac{l}{\delta \theta_k}
	\end{eqnarray}
	where $w$ is the distance between the driven wheels. Wheel encoders are used to measure $l_k$ and $r_k$ and therefore, $R_k$ and $\delta\theta_k$ can be determined at time step $k+1$. Then, the position of the robot is updated recursively as follows.
	\begin{equation}
	P_{k+1} = P_k +
	\begin{bmatrix}
	(R_k +w/2)(\sin(\theta_k+\delta\theta_k)-\sin\delta\theta_k)\\
	(R_k +w/2)(\cos\theta_k-\cos(\theta_k+\delta\theta_k)\\
	\delta\theta_k
	\end{bmatrix}
	\end{equation}
	where $P_k=(x_k,y_k,\theta_k)^T$.
	Due to wheel slipping, the recursive update of position using wheel encoders alone will accumulate errors \cite{thrun}. Hence, RTK and IMU (Inertial Measurement Unit) are used to correct the position and heading periodically.
\subsection{Sensor Fusion for Accurate Position Estimation}
	RTK-GPS positioning is expected to be very accurate despite its slow update rate (5Hz), and intermittent outages or drops in accuracy due to cloud cover. Hence, RTK-GPS is used to correct the wheel-encoder-based position update $(x_{k+1},y_{k+1})$. Similarly, the IMU sensor is used to correct the wheel encoder-based heading update $\theta_{k+1}$. Both RTK-GPS and IMU are absolute sensors unlike the wheel-encoder, which is a relative sensor.\par
\begin{figure}[ht]
	\centering
	\includegraphics[width=0.45\textwidth]{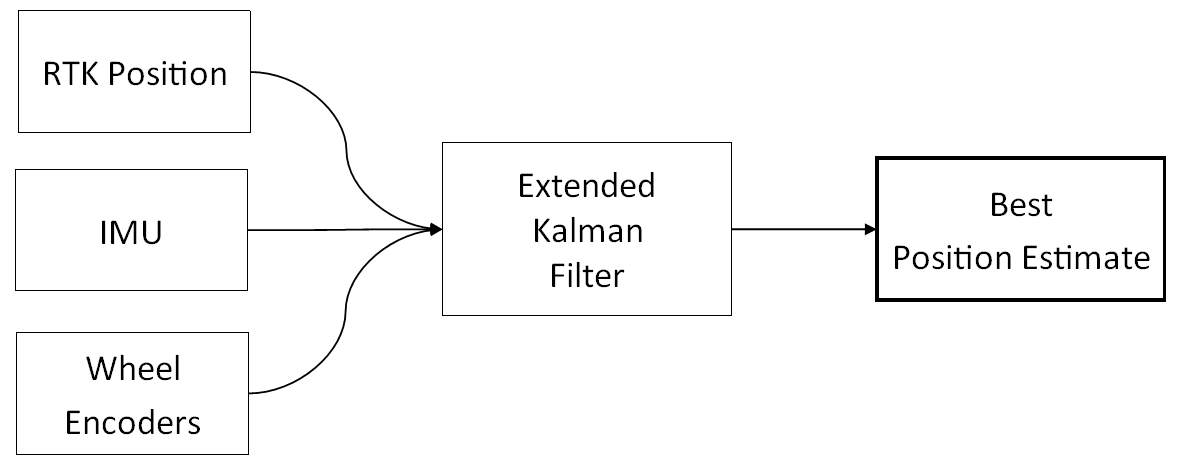}
	\caption{Extended Kalman filter for wheel-encoder, RTK-GPS, and IMU fusion for correct position/heading estimation}
	\label{best_pos}
	\end{figure}

	A ROS node receives GPS data from the GNSS and also the real-time GPS error from the RTK base station through an RF (radio frequency) link. Then, it determines the correct GPS location of the robot. The GPS coordinates are in LLH (Lattitude, longitude, height) format, hence it is transformed into Cartesian XYZ (Cartesian) format before combining with the wheel encoder data. This transformation is carried out using the ROS navsat transform node \cite{navsat_gps_link}. Then, the transformed RTK data, Wheel encoder data, and IMU data are fetched into the Extended Kalman Filter in the ROS navigation stack to get the accurate position and heading estimated. The extended Kalman filter and the position correction are shown in Figure \ref{best_pos}.\cite{ekf}
\subsection{Vision System}
	The robot's vision system, which consists of two Raspberry-PI cameras and an NVIDIA Jetson Nano, detects the plants on either side as it moves. The object detector used was the SSD-Mobilenet \cite{mobilenet}, which is computationally less complex so that a higher speed (FPS-frames per second) compared to other object detectors is possible. In order to make the input images compatible with the SSD-Mobilenet and to speed up inference, they are resized from 1080x720 to 224x224.
	In the SSD-Mobilenet model, non-maximum suppression is used to reduce the overlapping detections into a single bounding box with a 0.45 IoU (Intersection over Union) threshold. Transfer learning was used for a pre-trained SSD-Mobilenet model on the COCO dataset \cite{COCO} to train the SSD-MobileNet object detector model, and the batch size was set to 4. This model had an initial learning rate of 0.01 and a momentum of 0.9. The specifications of the vision system are as follows.
	The vision system was able to correctly detect objects at a substantially high speed of 19-24 FPS once implemented the Tensor RT default Jetson nano library. The machine vision system then split into two distinct ROS nodes and used those nodes to transfer the coordinate information that should be necessary for the calculations of the Plant Localization and Spray Control mechanism. In addition, the horizontal and vertical angles are calculated in relation to the camera's focal plane. Only the first instance of the detected plant passing the camera's center is used for the calculations. Lastly, the calculated angles will be sent to the plant localizing algorithm.
\section{Plant Localization and Spray Control}
\begin{figure*}[ht]
		\centering
		\includegraphics[width=14cm]{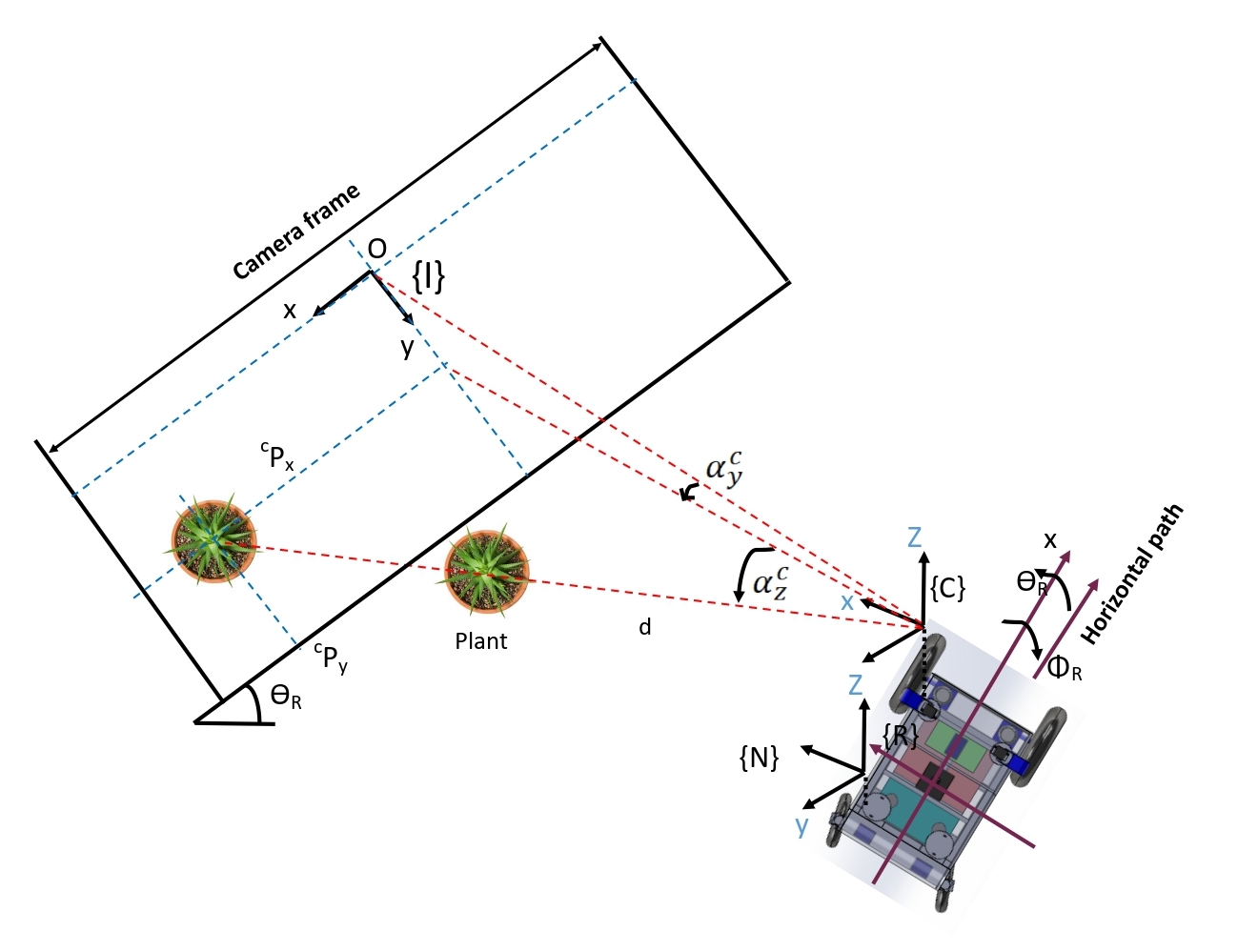}
		\caption{Coordinate system} \label{cord-system}
	\end{figure*}

\begin{figure}[ht]
		\centering
		\includegraphics[width=0.45\textwidth]{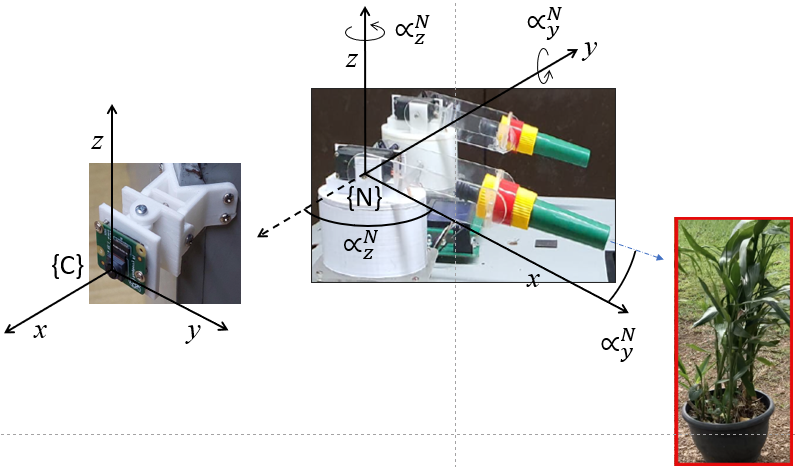}
		\caption{The camera, nozzle, nozzle angles and the plant} \label{nozzle_ang}
	\end{figure}

	When the vision system detects a plant in the image, the coordinates of the plant with respect to the camera frame $\{C\}$ are determined and then the homogeneous transformation of the nozzle with respect to the camera $^C_NH$ is used to determine the plant coordinates with respect to the nozzle $\{N\}$. Figure \ref{cord-system} shows an instance where the robot shows a pitch $\theta$ and a roll $\phi$ along with the camera $\{C\}$, spray nozzle $\{N\}$, image frame $\{I\}$ and a plant and its image.
	Using the pixel coordinates $(^cp_x,^cp_y)$ of the plant, the two spherical coordinates of the plant with respect to the camera can be determined using the camera parameters as follows:
	\begin{eqnarray}
		\alpha^c_y &=&f_y(^cp_y)\\
		\alpha^c_z &=&f_z(^cp_x)
	\end{eqnarray}
	where $f_y$ and $f_z$ are the functions that convert pixel length to angle along $y$ and $x$ axes of the image using the field of view and the pixel density of the camera. As shown in Fig.\ref{cord-system}, the homogeneous coordinates of the plant with respect to the camera are as follows.
	\begin{equation}
		^Cp=\begin{pmatrix}
			d\cos\alpha^c_z\\
			d\sin\alpha^c_z\\
			-d\cos\alpha^c_z\tan\alpha^c_y\\
			1
		\end{pmatrix}
	\end{equation}
	where $d$ is the distance to the plant.
	The nozzle frame at its home position is aligned with the camera frame $\{C\}$ and is located at $(20,0,10)^T$cm with respect to the camera. Hence, the homogeneous transformation matrix of the nozzle frame at its home position with respect to the camera is as follows.
	\begin{equation}
		^C_NH=\begin{bmatrix}
			1 & 0 & 0 & 20 \\
			0 & 1 & 0 & 0 \\
			0 & 0 & 1 & 10 \\
			0 & 0 & 0 & 1 
		\end{bmatrix}
	\end{equation}
	Then, the coordinates of the plant with respect to the nozzle frame are determined as follows.
	\begin{eqnarray}
		^Np &=&^C_NH^Cp\nonumber\\
		&=&\begin{pmatrix}
			d\cos\alpha^c_z+20 \\
			d\cos\alpha^c_z \\
			-d\cos\alpha^c_z\tan\alpha^c_y+10 \\
			1 
		\end{pmatrix}
	\end{eqnarray}
	Then, the two spherical coordinates (pan-tilt) of the nozzle to aim at the plant are as follows.
	\begin{eqnarray}
		\alpha^N_y&=& \tan^{-1}\left(\frac{-d\cos\alpha^c_z\tan\alpha^c_y+10}{d\cos\alpha^c_z+20}\right)\\
		\alpha^N_z&=& \tan^{-1}\left(\frac{d\cos\alpha^c_z}{d\cos\alpha^c_z+20}\right)
	\end{eqnarray}
	Figure \ref{nozzle_ang} shows the camera, nozzle, and the two angles of the nozzle to point it to the plant. 
	The two servo motors along $y$ and $z$ axes of the frame $\{N\}$ turn the nozzle around $y$ axis by $\alpha^N_y$ and around $z$ axis by $\alpha^N_z$ starting off the home position of $\{N\}$ and point the nozzle towards the plant.\par
	In continuous spraying the nozzle does not have to start off at the home position for each plant, instead, incremental angle adjustments from the previous plant to the next plant are determined as follows.
	\begin{eqnarray}
		\Delta\alpha^N_{y,n}&=& \alpha^N_{y,n} -\alpha^N_{y,n-1}\\
		\Delta\alpha^N_{z,n}&=& \alpha^N_{z,n}-\alpha^N_{z,n-1} 
	\end{eqnarray}
	where $n=1,2,3...$ is the plant index. The flow rate of the nozzle is 10l/min. Assuming each plant needs 200ml it takes 1.2s to spray a plant. During the spray time, the robot moves by 24cm given the average speed of 12m/s. In this research, this displacement is not considered an issue. 
	\section{Results}
The performance of subsystems has been tested as follows.
\subsection{RTK Positioning Accuracy}
A few known locations were used to test RTK accuracy. The robot was positioned at these locations and the real-time coordinates it received after implementing the correction from the RTK base station were observed. Table \ref{actual_observed} shows the converted position coordinates in XYZ format.
\renewcommand{\arraystretch}{1.8}{\begin{table}[H]
	\caption{Actual and observed coordinates}
	\label{actual_observed}
	\resizebox{\columnwidth}{!}{
		\begin{tabular}{|l|ll|ll|}
			\hline
			\multicolumn{1}{|c|}{\multirow{2}{*}{Direction}} & \multicolumn{2}{c|}{Point1} & \multicolumn{2}{c|}{Point2}\\
			\cline{2-5} 
			\multicolumn{1}{|c|}{}                           & \multicolumn{1}{c|}{\begin{tabular}[c]{@{}c@{}}Actual   \\ Coordinates\end{tabular}} & \multicolumn{1}{c|}{\begin{tabular}[c]{@{}c@{}}Observed   \\ Coordinates\end{tabular}} & \multicolumn{1}{c|}{\begin{tabular}[c]{@{}c@{}}Actual   \\ Coordinates\end{tabular}} & \multicolumn{1}{c|}{\begin{tabular}[c]{@{}c@{}}Observed   \\ Coordinates\end{tabular}} \\ \hline
			x(m)& \multicolumn{1}{l|}{1110825.867}     & 1110825.87085                            & \multicolumn{1}{l|}{1110706.361}         & 1110706.36502\\ \hline
			y(m) & \multicolumn{1}{l|}{6235329.584}    & 6235329.55216                            & \multicolumn{1}{l|}{6235347.832}         & 6235347.89016\\ \hline
			z(m)& \multicolumn{1}{l|}{750012.164}      & 750012.098407                            & \multicolumn{1}{l|}{750033.936}          & 750033.982406\\ \hline
	\end{tabular}}
\end{table}}
The error for the two points are 3.2cm and 5.8cm, respectively. Hence the mean error is 4.5cm.
\subsection{Path Tracking Accuracy}
\begin{figure}[H]
\centering
\includegraphics[width=0.35\textwidth]{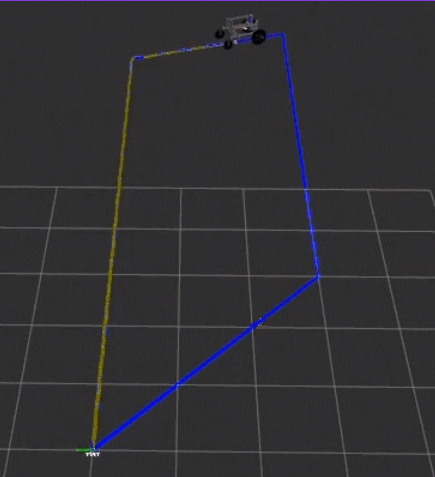}
\caption{Real-time feedback of the robot in the physical environment from RVIZ}
\label{real}
\end{figure}

Figure \ref{real} shows the motion in the RVIZ visualizer while the robot was moving on the ground. The mean and variance of the robot's path were calculated as 7cm and 25cm respectively. The fact that the motion shows a larger variance compared to the mean error is expected because of the free swinging of the front caster wheels. These two caster wheels quickly move when they go over ground imperfections, causing a higher variance in the path error.
\subsection{Plant Identification Accuracy}
\begin{figure}[H]
\centering
\includegraphics[width=0.4\textwidth]{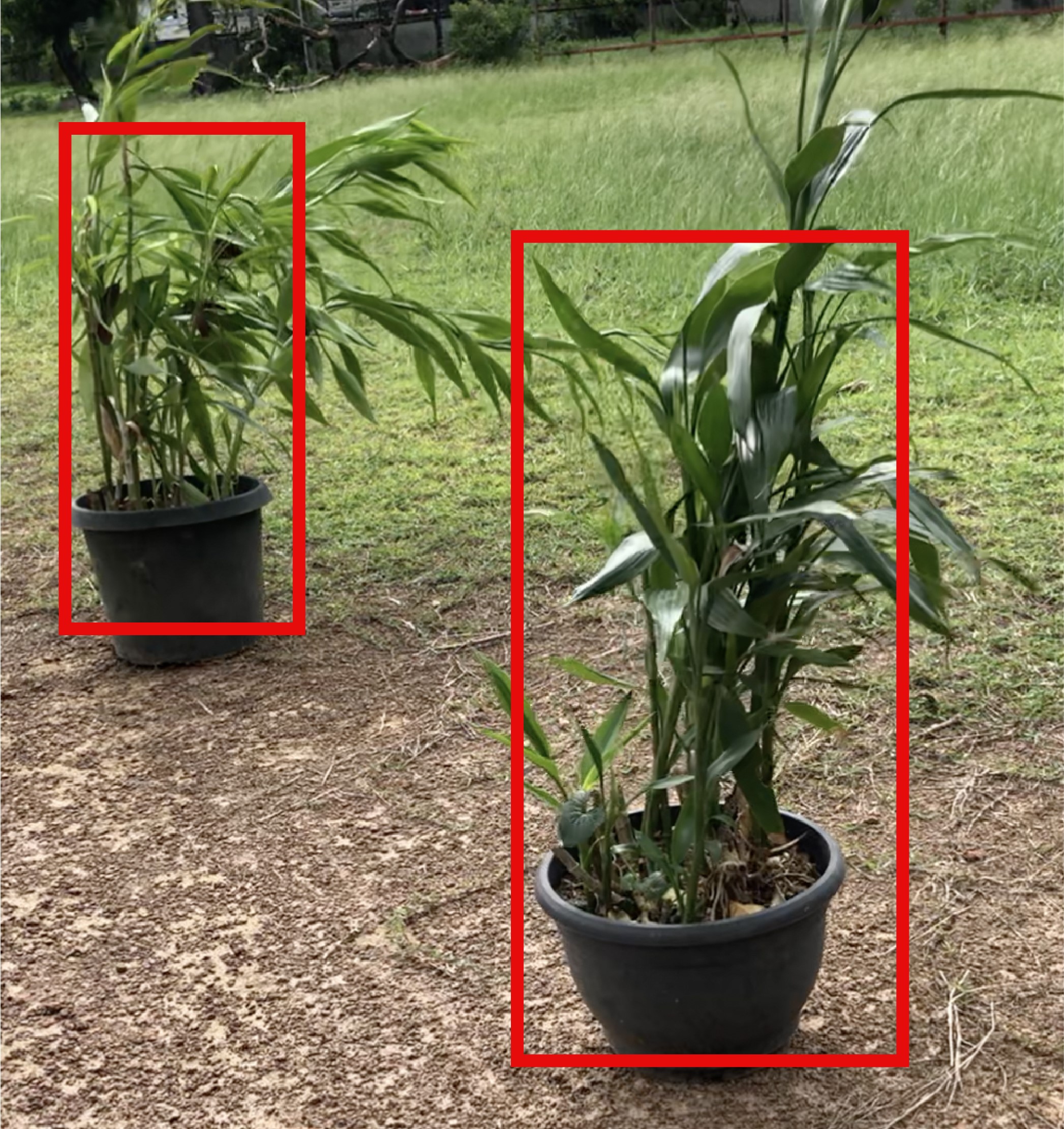}
\caption{Final detection results}
\label{mvr}
\end{figure}

The deep neural network was trained using the Tensorrt inference library and was able to maintain a real-time processing speed of 19~24 FPS with very high accuracy. Fig.\ref{mvr} shows plant identification by the trained deep neural network.
It can be verified that the SSD-MobileNet model performs better in terms of speed and accuracy. The bounding boxes of the identified plant are used in determining the pixel coordinates $^cp_x$ and $^cp_y$.
\section{Conclusion}
This research demonstrates a fully functional autonomous plant spraying robot. The navigation system was designed using RTK-GPS, wheel encoders, and an IMU, whereas the spraying system works on a trained deep neural network for plant identification and a pan-tilt robot mechanism to control the spray nozzle direction towards the selected plant. Accurate path following, plant detection, and accurate spraying have been demonstrated.\par
This robot sprays the right amount of agro-chemicals onto the plant, and its operational time can be extended by replacing the battery with a small gasoline-powered engine. Hence, the outcome of this research has the potential of becoming a smart agri-technology in the future.
\bibliographystyle{IEEEtran}
\bibliography{reference}

\begin{thebibliography}{10}
\providecommand{\url}[1]{#1}
\csname url@samestyle\endcsname
\providecommand{\newblock}{\relax}
\providecommand{\bibinfo}[2]{#2}
\providecommand{\BIBentrySTDinterwordspacing}{\spaceskip=0pt\relax}
\providecommand{\BIBentryALTinterwordstretchfactor}{4}
\providecommand{\BIBentryALTinterwordspacing}{\spaceskip=\fontdimen2\font plus
\BIBentryALTinterwordstretchfactor\fontdimen3\font minus
  \fontdimen4\font\relax}
\providecommand{\BIBforeignlanguage}[2]{{%
\expandafter\ifx\csname l@#1\endcsname\relax
\typeout{** WARNING: IEEEtran.bst: No hyphenation pattern has been}%
\typeout{** loaded for the language `#1'. Using the pattern for}%
\typeout{** the default language instead.}%
\else
\language=\csname l@#1\endcsname
\fi
#2}}
\providecommand{\BIBdecl}{\relax}
\BIBdecl

\bibitem{kusumam20173d}
K.~Kusumam, T.~Krajn{\'\i}k, S.~Pearson, T.~Duckett, and G.~Cielniak,
  ``3d-vision based detection, localization, and sizing of broccoli heads in
  the field,'' \emph{Journal of Field Robotics}, vol.~34, no.~8, pp.
  1505--1518, 2017.

\bibitem{nakarmi2014within}
A.~D. Nakarmi and L.~Tang, ``Within-row spacing sensing of maize plants using
  3d computer vision,'' \emph{Biosystems engineering}, vol. 125, pp. 54--64,
  2014.

\bibitem{mccool2018efficacy}
C.~McCool, J.~Beattie, J.~Firn, C.~Lehnert, J.~Kulk, O.~Bawden, R.~Russell, and
  T.~Perez, ``Efficacy of mechanical weeding tools: A study into alternative
  weed management strategies enabled by robotics,'' \emph{IEEE Robotics and
  Automation Letters}, vol.~3, no.~2, pp. 1184--1190, 2018.

\bibitem{wu2019design}
X.~Wu, S.~Aravecchia, and C.~Pradalier, ``Design and implementation of computer
  vision based in-row weeding system,'' in \emph{2019 International Conference
  on Robotics and Automation (ICRA)}.\hskip 1em plus 0.5em minus 0.4em\relax
  IEEE, 2019, pp. 4218--4224.

\bibitem{rtk_link}
\BIBentryALTinterwordspacing
``Rtk gps: Understanding real-time kinematic gps technology.'' [Online].
  Available:
  \url{https://globalgpssystems.com/gnss/rtk-gps-understanding-real-time-kinematic-gps-technology/}
\BIBentrySTDinterwordspacing

\bibitem{misson_planner}
``Mission planner overview.''

\bibitem{Olson2011AprilTagAR}
E.~Olson, ``Apriltag: A robust and flexible visual fiducial system,''
  \emph{2011 IEEE International Conference on Robotics and Automation}, pp.
  3400--3407, 2011.

\bibitem{88147}
J.~Leonard and H.~Durrant-Whyte, ``Mobile robot localization by tracking
  geometric beacons,'' \emph{IEEE Transactions on Robotics and Automation},
  vol.~7, no.~3, pp. 376--382, 1991.

\bibitem{bell}
T.~Bell, ``Automatic tractor guidance using carrier-phase differential gps,''
  \emph{Computers and Electronics in Agriculture}, vol.~25, pp. 53--66, 01
  2000.

\bibitem{thuilot}
B.~Thuilot, C.~Cariou, P.~Martinet, and M.~Berducat, ``Automatic guidance of a
  farm tractor relying on a single cp-dgps,'' \emph{Autonomous Robots},
  vol.~13, pp. 53--71, 07 2002.

\bibitem{Underwood2015RealtimeTD}
J.~P. Underwood, M.~Calleija, Z.~Taylor, C.~Hung, J.~I. Nieto, R.~C. Fitch, and
  S.~Sukkarieh, ``Real-time target detection and steerable spray for vegetable
  crops,'' 2015.

\bibitem{Imperoli2018AnEM}
M.~Imperoli, C.~Potena, D.~Nardi, G.~Grisetti, and A.~Pretto, ``An effective
  multi-cue positioning system for agricultural robotics,'' \emph{IEEE Robotics
  and Automation Letters}, vol.~3, pp. 3685--3692, 2018.

\bibitem{Dong20174DCM}
J.~Dong, J.~G. Burnham, B.~Boots, G.~C. Rains, and F.~Dellaert, ``4d crop
  monitoring: Spatio-temporal reconstruction for agriculture,'' \emph{2017 IEEE
  International Conference on Robotics and Automation (ICRA)}, pp. 3878--3885,
  2017.

\bibitem{Chebrolu2019RobotLB}
N.~Chebrolu, P.~Lottes, T.~L{\"a}be, and C.~Stachniss, ``Robot localization
  based on aerial images for precision agriculture tasks in crop fields,''
  \emph{2019 International Conference on Robotics and Automation (ICRA)}, pp.
  1787--1793, 2019.

\bibitem{Billingsley1997TheSD}
J.~Billingsley and M.~Schoenfisch, ``The successful development of a vision
  guidance system for agriculture,'' \emph{Computers and Electronics in
  Agriculture}, vol.~16, pp. 147--163, 1997.

\bibitem{astrand}
B.~Åstrand and A.-J. Baerveldt, ``A vision based row-following system for
  agricultural field machinery,'' \emph{Mechatronics}, vol.~15, pp. 251--269,
  03 2005.

\bibitem{143350}
B.~Espiau, F.~Chaumette, and P.~Rives, ``A new approach to visual servoing in
  robotics,'' \emph{IEEE Transactions on Robotics and Automation}, vol.~8,
  no.~3, pp. 313--326, 1992.

\bibitem{Cherubini2008AnIV}
A.~Cherubini, F.~Chaumette, and G.~Oriolo, ``An image-based visual servoing
  scheme for following paths with nonholonomic mobile robots,'' \emph{2008 10th
  International Conference on Control, Automation, Robotics and Vision}, pp.
  108--113, 2008.

\bibitem{768184}
Y.~Ma, J.~Kosecka, and S.~Sastry, ``Vision guided navigation for a nonholonomic
  mobile robot,'' \emph{IEEE Transactions on Robotics and Automation}, vol.~15,
  no.~3, pp. 521--536, 1999.

\bibitem{this}
A.~Ahmadi, L.~Nardi, N.~Chebrolu, and C.~Stachniss, ``Visual servoing-based
  navigation for monitoring row-crop fields,'' 09 2019.

\bibitem{park2004new}
S.~Park, J.~Deyst, and J.~How, ``A new nonlinear guidance logic for trajectory
  tracking,'' in \emph{AIAA guidance, navigation, and control conference and
  exhibit}, 2004, p. 4900.

\bibitem{thrun}
\BIBentryALTinterwordspacing
S.~Thrun, W.~Burgard, and D.~Fox, \emph{Probabilistic Robotics}, ser.
  Intelligent Robotics and Autonomous Agents series.\hskip 1em plus 0.5em minus
  0.4em\relax MIT Press, 2005. [Online]. Available:
  \url{https://books.google.lk/books?id=k_yOQgAACAAJ}
\BIBentrySTDinterwordspacing

\bibitem{navsat_gps_link}
\BIBentryALTinterwordspacing
T.~Moore, ``Integrating gps data,'' 2016. [Online]. Available:
  \url{https://docs.ros.org/en/melodic/api/robot_localization/html/integrating_gps.html}
\BIBentrySTDinterwordspacing

\bibitem{ekf}
\BIBentryALTinterwordspacing
------, ``State estimation nodes,'' 2016. [Online]. Available:
  \url{https://docs.ros.org/en/melodic/api/robot_localization/html/state_estimation_nodes.html#ekf-localization-node}
\BIBentrySTDinterwordspacing

\bibitem{mobilenet}
M.~Sandler, A.~Howard, M.~Zhu, A.~Zhmoginov, and L.-C. Chen, ``Mobilenetv2:
  Inverted residuals and linear bottlenecks,'' in \emph{Proceedings of the IEEE
  conference on computer vision and pattern recognition}, 2018, pp. 4510--4520.

\bibitem{COCO}
\BIBentryALTinterwordspacing
T.-Y. Lin, M.~Maire, S.~Belongie, L.~Bourdev, R.~Girshick, J.~Hays, P.~Perona,
  D.~Ramanan, C.~L. Zitnick, and P.~Dollár, ``Microsoft coco: Common objects
  in context,'' 2014. [Online]. Available:
  \url{https://arxiv.org/abs/1405.0312}
\BIBentrySTDinterwordspacing

\end{thebibliography}
\end{document}